\documentclass[conference]{IEEEtran}
\renewcommand{\baselinestretch}{1.0}
\IEEEoverridecommandlockouts

\usepackage{cite}
\usepackage{amsmath,amssymb,amsfonts} %Mathematical Symbol,Equation
\usepackage{algpseudocode}
\usepackage[linesnumbered,ruled,vlined]{algorithm2e}
\usepackage{graphicx} %Adding picture and figure
\usepackage{float} %Figure Alignment
\usepackage{textcomp} %Text formatting
\usepackage{eso-pic}
\usepackage{xcolor}
\usepackage{comment}
\usepackage{csquotes}
\def\BibTeX{{\rm B\kern-.05em{\sc i\kern-.025em b}\kern-.08em
    T\kern-.1667em\lower.7ex\hbox{E}\kern-.125emX}}
    
\makeatletter
\def\ps@IEEEtitlepagestyle{%
  \def\@oddfoot{\mycopyrightnotice}%
  \def\@evenfoot{}%
}
\def\mycopyrightnotice{%
{\footnotesize 979-8-3503-8577-9/24/\$31.00 \textcopyright2024 IEEE\hfill}% <--- Change here
  \gdef\mycopyrightnotice{}% just in case
}

\usepackage[textsize=footnotesize, textwidth=1.2cm]{todonotes}

\usepackage{eso-pic}
\usepackage{fancyhdr}
\begin{document}

%\title{A Remote Control Painting System for Outside Walls of High Rise Building}
\title{A Remote Control Painting System for Exterior Walls of High-Rise Buildings through Robotic System}

%\title{A Robotic-based Remote Control Painting System for Exterior Walls of High-Rise Buildings}

%\begin{comment}
\author{\IEEEauthorblockN{Diganta Das\IEEEauthorrefmark{1}, Dipanjali Kundu\IEEEauthorrefmark{2}, Anichur Rahman *\IEEEauthorrefmark{2}, Muaz Rahman\IEEEauthorrefmark{2} and Sadia Sazzad \IEEEauthorrefmark{2}}

\IEEEauthorblockA{\textit{Department of Statistics},
\textit{Jahangirnagar University, Dhaka}\\
\textit{Department of CSE \& EEE, National Institute of Textile Engineering and Research (NITER)} \\
%\textit{Senior Specialist, Digital Security \& Digital Diplomacy, ICT Division, Agargaon, Dhaka-1207} \\
digantadas132@gmail.com\IEEEauthorrefmark{1}, 
dkundu@niter.edu.bd\IEEEauthorrefmark{2},
anis\_cse@niter.edu.bd\IEEEauthorrefmark{2},
muaz@niter.edu.bd\IEEEauthorrefmark{2}, and
ssazzad@niter.edu.bd\IEEEauthorrefmark{2}}}
%\end{comment}

\maketitle

\maketitle
%\conf{2024 6th International Conference on Electrical Engineering and Information \& Communication Technology (ICEEICT)\\
%Military Institute of Science and Technology (MIST), Dhaka-1216, Bangladesh}

%%%%%%%%%%%%%%%%%%%%%%%%%%%%%%%%%%%%%%%%%%%%%%%%%%%%%%%%%%%%%%%%%%%%%%%%%%%%%%%%

\begin{abstract}
Exterior painting of high-rise buildings is a challenging task. In our country, as well as in other countries of the world, this task is accomplished manually, which is risky 
and life-threatening for the workers. Researchers and industry experts are trying to find an automatic and robotic solution for the exterior painting of high-rise building walls. In this paper, we propose a solution to this problem. We design and implement a prototype for automatically painting the building walls' exteriors. A spray mechanism was introduced in the prototype that can move in four different directions (up-down and left-right). All the movements are achieved by using microcontroller-operated servo motors. Further, these components create a scope to upgrade the proposed remote-controlled system to a robotic system in the future. In the presented system, all the operations are controlled remotely from a smartphone interface. Bluetooth technology is used for remote communications. It is expected that the suggested system will improve productivity with better workplace safety. 
\end{abstract}

\vspace{2mm}

\section{Introduction}
Building construction is one of the major industries in not only Bangladesh but also around the world. In this modern era, the construction industry is also growing rapidly \cite{rahman2021smartblock, kumtole2022automatic}. The global building and construction sector stands as a major industry, experiencing rapid growth in our fast-paced world. However, a significant challenge it faces is the scarcity of labor. This scarcity primarily stems from the demanding nature of construction work, especially in scenarios involving tall buildings or high-risk sites within urban areas \cite{10103295}. Several factors contribute to this insufficient labor force. One crucial element is the perception surrounding these jobs—improvements in education have led some to view construction work as less prestigious compared to other professions. This shift in perception impacts the industry's ability to attract a robust workforce. Acknowledging the labor-intensive and often hazardous nature of construction, there has been a substantial realization of the importance of integrating construction robotics. The emergence of robotics and automation in construction traces back to the early 90s, initially aimed at optimizing equipment operations, bolstering safety measures, improving workspace awareness, and ensuring a higher-quality environment for building occupants \cite{rahman2022sdn}.\vspace{2mm}

Since then, there has been a notable surge in advancements in robotics and automation in the construction industry. This evolution continues to reshape the industry, offering innovative solutions to enhance efficiency, safety, and overall productivity while mitigating the constraints posed by a labor shortage. 
%Statistics shows that Bangladesh will need to construct approximately 4 million new houses. More than 4 million apartments in Bangladesh are built per year, having an average painting area of 40 million square meters (based on an average 4000 m2 painting area).Expected population will increase in the future and the surface area where paint is needed is more due to the renovation work. Like all other countries in the world, the house planning industry plays many important roles in both context of Bangladesh’s economy and serves the fundamental human right to protection. 
As a consequence, introducing a remote control painting system for high-rise buildings revolutionizes the conventional methods of exterior maintenance. This system poses challenges that human painters will hardly achieve in the next decade. Therefore, it is necessary to develop such a machine capable of performing the purpose of minimizing human effort and improving the quality of the painting like what we are thinking. The need for a system-target robot is both clear and strong. Service robot development has become very popular recently because robots relax people from boring and dangerous jobs \cite{yahya2020automatic}. This innovative system embodies cutting-edge technology, combining robotics and precision engineering to streamline and elevate the process of painting expansive vertical surfaces. While a remote control painting system for high-rise buildings offers numerous advantages, it also faces several challenges and potential issues. Adaptability to surfaces, precision and consistency, weather sensitivity, power and connectivity are the major challenges that were found in implementing this technology. Additionally, the process is becoming costly day by day \cite{patil2020man}. In response to the challenges posed by traditional manual painting methods for high-rise structures, our remote-control painting system emerges as a sophisticated solution. Technology employed in electronics, computers, even mechanics, and hydraulics can all be measured as a fraction of the robotics history. Robots can also be used in the painting process of a building. A wall painting robot is a machine that can be used to paint the walls of houses, offices, hospitals, etc. By integrating robotic capabilities with remote control functionality, this system aims to transform the efficiency, safety, and precision of exterior wall painting on tall buildings \cite{megalingam2020autonomous}.   \vspace{2mm}
 
Moreover, maintaining high-rise buildings in contemporary urban landscapes is an important and difficult exercise. Among the different building maintenance aspects, painting exterior walls stands out owing to its intricacy and significant implications for the aesthetics, longevity, and value of a structure. Typically, manual labor is used when painting skyscrapers, which is not only dangerous but also time-consuming and expensive \cite{rahman2020distblockbuilding}. These difficulties are addressed through the development of a remote control painting system (RCPS) for the external walls of high-rise buildings that offers an innovative solution that merges safety, productivity, and technological advantage. Thus, the main motivation of this system is to build a system that can paint the walls automatically in a once-through fashion vertically from starting point to end, be stable and not turn over during painting, and paint the side walls and sunroof. This robot will save human efforts and improve throughput, reduce the environmental risks to human lives, and overall raise the quality of work. However, it's important to note that creating such a system would involve significant technological development, rigorous testing, and collaboration with experts in robotics, construction, and safety regulations. Additionally, gaining acceptance and clearance for using such a system on actual high-rise buildings would involve navigating various logistical and legal considerations.
% \textbf{Benefits:} Our remote control painting system represents a leap forward in the realm of building maintenance, offering an innovative and efficient alternative to traditional methods. By combining technological prowess with a commitment to safety and efficiency, this system aims to redefine the standards for exterior wall painting on high-rise buildings.
%\begin{itemize} 
%\item {\textbf{Safety:} Reduce risks associated with manual painting at heights.}
%\item {\textbf{Efficiency:} Increase speed and accuracy of painting large surfaces.}
%\item {\textbf{Cost-Effectiveness:} Minimize paint wastage and reduce labor costs over time.}
%\end{itemize}
The main contributions of this paper are--
 \begin{itemize}
 \item {\textbf{Robotic Painting Apparatus:} To make a machine that is structurally simple to enable easy mounting and also safe.}\vspace{2mm}
 %Design a robotic arm or system that can maneuver across the walls of high-rise buildings. Ensure it's equipped with painting modules, reservoirs for paint, and a mechanism for controlled and even paint application.}
 
\item {\textbf{Remote Control Interface:} To paint the outside wall of a high-rise building in a single color.}\vspace{2mm}
%Develop a user-friendly interface allowing operators to control the robot remotely.
%Include features for adjusting painting speed, paint quantity, and precision controls.}

\item{\textbf{Safety Mechanisms:} To paint the outside wall from top to bottom and from left to right with safety mechanisms.
%Implement sensors to detect obstacles, changing weather conditions, and ensure the safety of workers and pedestrians below.Emergency stop mechanisms in case of malfunctions or sudden hazards.}
%\item {\textbf{Stability and Adaptability:} Ensure the system can adapt to different surface textures, angles, and heights typical of high-rise buildings. Stability mechanisms to handle wind, vibrations, and other external factors.}
%\item {\textbf{Power and Connectivity:} Power source considerations, which might involve a combination of batteries, rechargeable systems, or tethered power sources. Reliable communication systems for remote operation and data transmission.}
%\item {\textbf{Painting Material and Efficiency:} Develop or choose painting materials that adhere well to various surfaces, are weather-resistant, and require minimal reapplications. Design the system for efficient paint usage to minimize waste and ensure cost effectiveness.
}
\end{itemize}\vspace{2mm}

Therefore, there will be no need to make a bamboo stage around the building. Our robot will automatically paint the whole building by shifting walls one after another, saving a lot of time.
% Development Process: Research and Design Phase: Understand the different types of building materials and surfaces the system will encounter. Design prototypes, considering various climatic conditions and potential challenges. Testing and Refinement: Conduct extensive testing in controlled environments to ensure stability, efficiency, and safety. Gather feedback and refine the system based on real-world simulations. Regulatory Compliance: Ensure compliance with safety and regulatory standards for construction and painting activities on high-rise buildings. Deployment and Maintenance: Train operators for system use and maintenance. Establish protocols for regular inspections, maintenance, and updates.

\vspace{2mm}
\textcolor{black}{\textbf{Organization:} The rest of the article is arranged as follows: relevant concepts of automated wall painting robots and conventional systems are presented in Section \ref{RW}. In section \ref{PM}, the proposed methodology is discussed with motivation, followed by a brief result analysis and discussion in Section \ref{RD}. Finally, the conclusion with future scope is discussed in Section \ref{sec:conclu}}.

% \paragraph*{\textbf{Overview of Existing Survey Papers}}
\section{Related Works} \label{RW}
The traditional method of external wall paintings involved human labor in a hazardous environment.  Automated wall painting will considerably reduce human labor and mitigate the deadly accidents that might occur. This will also ensure an efficient method of reaching the desired output in a short time frame. The risky nature of such work put researchers into motion towards a more effective and safe means to paint the exterior of high-rise buildings. Researchers have discussed several works on the deployment of wall-painting robots.

\vspace{2mm}
The work presented in \cite{takeno1990apparatus} presents a robot used for external wall paintings through the use of an apparatus. The mechanism involved advancing and retracting techniques and the ability to maneuver horizontally. The work in \cite{thakar2014review, rahman2024blocksd} outlines the necessity of automation for coating or painting purposes. This method will reduce time and provide a higher degree of accuracy. Researchers in \cite{yeom2017performance} focused on exterior wall painting in South Korea. The authors proposed a Gondola-type Exterior Wall Painting Robot (GEWPro) and calculated its performance and economic analysis. The model achieved considerable success over conventional or human-labor-based methods of exterior wall paintings. The model developed in \cite{lim2019system} utilized a six-axis robotic arm which is placed on a small-sized gondola. The mechanism further deploys ultrasonic distance sensors, which calculate the distance between the device and the wall surface. The effect of wind load on robots used for exterior wall painting is studied in \cite{cho2015wind}. The developed prototype is put through computational fluid dynamics simulation to verify its stability against the high wind at greater heights. The techno-economic aspects of utilizing a wall painting robot for the exterior of a building are studied extensively in \cite{kim2006conceptual}. Researchers in \cite{teoh2011paintbot} developed a prototype robot used for wall painting in indoor and outdoor surroundings. Similar to external wall painting robots, several researchers have carried out research into indoor wall painting prototypes. Authors in \cite{keerthanaa2013automatic}, detail the functioning of an internal wall-painting robot. The concept presented is used for a residential dwelling where a robot with a roller paints the wall. The proposed model possesses the ability to accurately maintain a position through an IR transmitter and receiver. Design considerations for an indoor wall painting robot is analyzed in \cite{abdellatif2013system}. Indoor wall painting using a sensor-based prototype is detailed in \cite{mukundan2017automatic}. A computer-aided design model for a robot capable of interior wall painting is discussed by authors in \cite{sorour2015robopainter}. Research into unmanned aerial vehicles used for wall painting in an industrial setting is discussed by authors in \cite{vempati2018paintcopter}. Again, a drone mechanism is deployed in \cite{lai2018automatic} to paint a structure. 
\vspace{2mm}

In summary, we investigated multiple prospective studies and discovered numerous remarkable outcomes from various scholars. Based on this, the proposed remote-control painting system employs robotics, remote control, and cutting-edge materials to address concerns of risk, effectiveness, and workmanship in exterior wall painting of high-rise buildings. The system aims to improve productivity and workplace safety. Additionally, the basic prototype has limitations in handling obstructions like sunroofs or arches, which need further development. 

% \end{new}

% \paragraph*{\textbf{Methodology}}
\section{Proposed Methodology for Remote Control Painting System for High-Rise Building Exterior Walls}
\label{PM}
In this section, we discuss the machine's components and how a Bluetooth device may operate it from a long distance.

\begin{figure*}[ht!]
\centering
\includegraphics[scale=0.62]{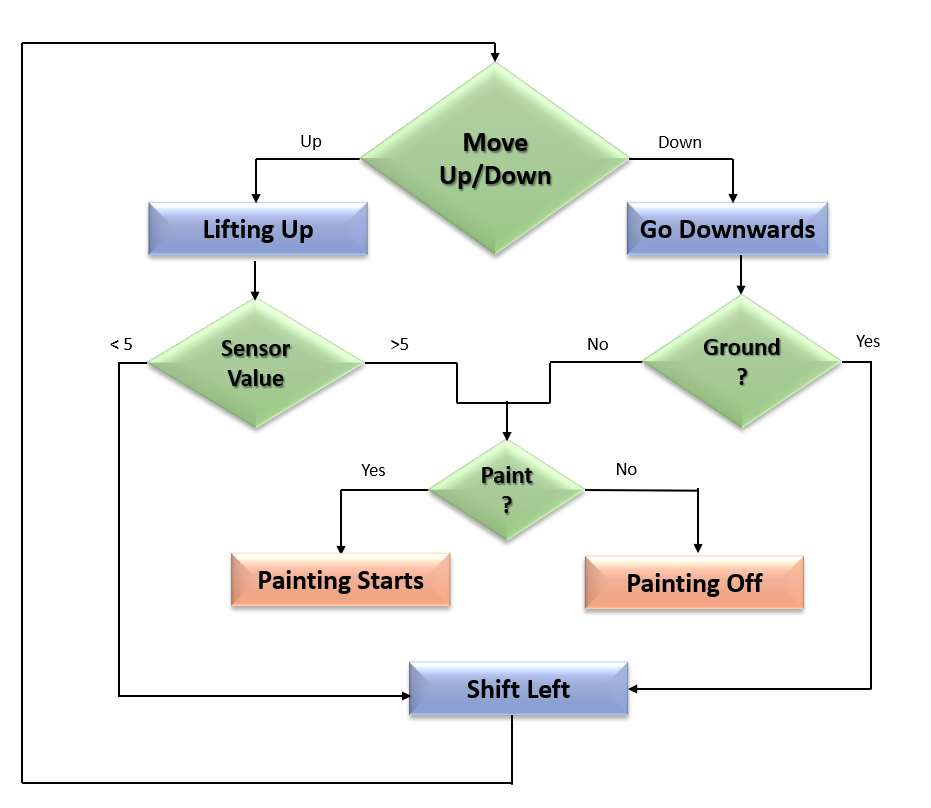}
\caption{Proposed Flow-diagram for Wall Painting Robot}
\label{fig:Fig1}%
\end{figure*}

\subsection{Overviews of the Proposed System}
The primary goal of this system is to create an automated system that can paint walls vertically from top to bottom in a single pass. While painting the walls and ceilings, maintain equilibrium and avoid tipping over. This robot would reduce the need for human labor, increase throughput, lessen the risk to human health from the environment, and generally provide higher-quality work. The painters dangle from the building's roof by a rope or construct a bamboo platform around the structure and proceed with painting. It's a hazardous way to paint a structure. The identical task could potentially be completed by a robot but in a different way. Therefore, the goal of this project is to construct a robot that can autonomously paint the entire structure. The appropriate reason for applying automation is to boost productivity and quality beyond what is possible with current human labor levels to realize economies of scale and predictable quality levels. The inappropriate application of automation, which arises most often, is a tendency to eliminate or replace human labor simply because the correct application of automation can net as much as 3 to 4 times the original output with no increase in current human labor costs \cite{islam2021blockchain}.

\subsection{Procedure of Proposed Model}

Fig. \ref{fig:Fig1} presents the Proposed model for the wall painting robot. 

\begin{table*}[]
\centering
\caption{Major Components Used with their Purpose}
\begin{tabular}{ll}
\hline
\multicolumn{1}{|l|}{Component Name}& \multicolumn{1}{l|}{Purpose} \\ \hline
\multicolumn{1}{|l|}{Arduino Mega 2560} &
  \multicolumn{1}{l|}{Main component (Considered to be  the brain of the robot)}\\ \hline
\multicolumn{1}{|l|}{L298N}  & \multicolumn{1}{l|}{Moto controlling Mechanism}                     \\ \hline
\multicolumn{1}{|l|}{Stepper Motor} &
  \multicolumn{1}{l|}{For the easy movement of the Robotic arm to any degree} \\ \hline
\multicolumn{1}{|l|}{DC Motor}& \multicolumn{1}{l|}{For the movement of the machine along the wall} \\ \hline
\multicolumn{1}{|l|}{Servo Motor}            & \multicolumn{1}{l|}{Controlling of the arm for the paiting}         \\ \hline
\multicolumn{1}{|l|}{Ultrasonic Sensor (HCSR04)} &
  \multicolumn{1}{l|}{For the detection of wall and the distances from the wall} \\ \hline
\multicolumn{1}{|l|}{Limit Switch} & \multicolumn{1}{l|}{For determining the movement limit}             \\ \hline
\multicolumn{1}{|l|}{Bluetooth Module(HC05)} & \multicolumn{1}{l|}{For wireless connectivity} \\ \hline
\multicolumn{1}{|l|}{Power Supply}           & \multicolumn{1}{l|}{For the connection to run the machine} \\ \hline             
\end{tabular}

\label{tab:com}
\end{table*}

\subsubsection{Hardware}
\begin{itemize}
    \item \textbf{Arduino Mega 2560: }It is an ATmega2560-based microcontroller board. Includes 16 analog inputs, 4 serial ports (UART), 16 MHz crystal oscillator, 54 digital ip/output pins, of which 14 are utilized as PWM outputs, a connector for USB, a power jack, an ICSP header, along with a reset button. This board assists in avoiding the mixing of SI and CGS units, such as those currently in amperes. Hence it is necessary to clearly state the units for each quantity used \cite{elassal2024low, rahman2023impacts}.

    \item \textbf{Ultrasonic Sensors:} This sensor uses ultrasonic waves to measure distance. An ultrasonic wave is emitted by the sensor head, which then receives the wave reflected from the target. To calculate the target’s distance, this sensor uses a time interval between the emission and reception \cite{rani2024innovations}. Fig. \ref{fig:f2} presents the mechanism that the ultrasonic sensor follows. 
    
\begin{figure}[h]
    \centering
    \includegraphics[scale=0.90]{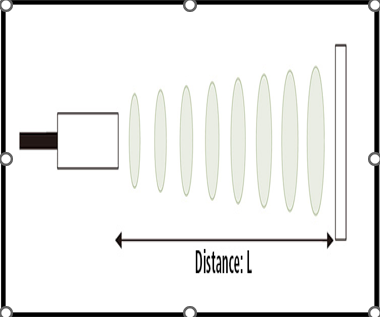}
   \caption{\textcolor{black}{Movement of Sound}}
    \label{fig:f2}
\end{figure}
    
    \item \textbf{Servo Motor: }Servo motors (SM)  which is considered to be a widely used motor in various applications based on robotics. The main advantage of using an SM is its small size, and despite its size, it has a lot of power and, on the other, uses less energy. It includes a movable shaft, which is often equipped with a gear. The servo electronics are integrated directly into its motor unit \cite{pillai2024robopaint}. The motor is operated by an electronic signal, which determines the movement of the shaft. Fig. \ref{fig:f3} presents the different pulses of the servo motor. 
\end{itemize}

     \begin{figure}[h]
    \centering
    \includegraphics[scale=0.70]{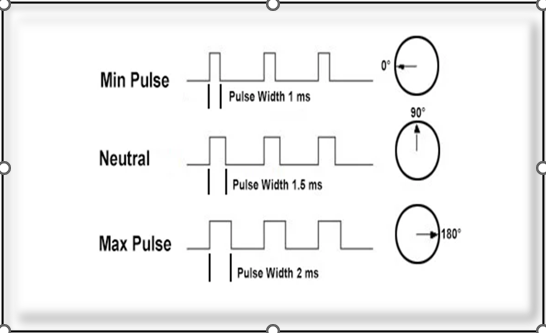}
   \caption{\textcolor{black}{Servo Motor}}
    \label{fig:f3}
\end{figure}

\subsubsection{Spining Mechanism}
We have to build a roller that can spin a rope to construct this machine. The machine's four corners should then have four ropes passing through them and standing on hanging stands. After that, attach the spinner to the motor, which enables it to carry out the assigned work. It's time to install the system after it has been prepared. The robot will first be suspended from the top of the wall, and it will begin to function as soon as the Arduino receives electricity. For this reason, as stated below, we have recommended the following methodology: 
\subsubsection{Startup}
Upon powering on, an Arduino script will begin to execute. An Arduino script is a set of instructions that specify the actions the Arduino should take upon starting. The software will launch after the bash script is executed. The painting technique and up-and-down movement are included in the program. It will move to the left when painting is finished.
\subsubsection{Up and Down Process}
The Arduino software is as basic as the C++ programming language. Each line in this case is executable. To run the devices outside of the Arduino with certain functionality, an additional component is installed. The write function is used to drive motors and do similar tasks to other functions.
\subsubsection{Painting Process}
The workings of a painting spray are common knowledge. All that's needed to spray paint it enclosed is pressure. Here, we paint with black paint. To push the bottle and begin painting, we also employ a mechanism with a servo motor that can apply pressure at the top of the bottle. We have included a photo of the spray paint mechanism here. 
\subsubsection{Shifting Process}
After painting one side entirely, the shifting process will begin. It's what the sound sensor will do. The sound sensor detects the machine's startup and initiates the motor to distort the entire apparatus. It will be twisted until it receives the command to descend. In Algorithm \ref{dposalgo}, the process of each robot's functionality is presented briefly.
\begin{algorithm}
\caption{Proposed Algorithm for Exterior Wall-Painting Robot Movement}
\label{dposalgo}
\SetAlgoLined
\DontPrintSemicolon

\SetKwInOut{Input}{Input}
\SetKwInOut{Output}{Output}
\SetKwInOut{Define}{Define}
\SetKwInOut{Initialize}{Initialize}

\Define{Step\_x, Step\_y}
\Initialize{StepperMotors(), ServoMotor(), Sensors(), PaintMove(), Paint()}

\SetKwFunction{FMain}{Main}
\SetKwFunction{FMoveToPosition}{MoveToPosition}
\SetKwFunction{FPaint}{Paint}
\SetKwFunction{FObstacleDetect}{ObstacleDetect}

\SetKwProg{Fn}{Function}{}{}

\Fn{\FMain{step\_x, step\_y}}{
    \While{ServoMotor() == True}{
        Connect to pin 3
    }
    \While{StepperMotors() == True}{
        Set speed of step\_x = 60 and step\_y = 60
    }
}

\Fn{\FMoveToPosition}{
    \For{position in (x, y)}{
        \FMoveToPosition{x, y}
        \FPaint{}
    }
}

\Fn{\FPaint}{
    Start the paint process by setting Servo position to 90-degree angle\;
    Delay for 500\;
    Set servo position to 0-degree angle for stop
}

\Fn{\FObstacleDetect}{
    \If{(Ultrasonic sensor value detects obstacle)}{
        Stop Paint\;
        Adjust path for the robot
    }
}

\end{algorithm}

\begin{comment}
 \begin{algorithm}
\label{dposalgo}
\SetAlgoLined
\caption{Proposed Algorithm for Exterior Wall-Painting Robot Movement}

\SetKwInOut{define}{define}
\SetKwInOut{initialize}{initialize}
  \SetKwFunction{Fmain}{pin\_{setup}}
  \SetKwFunction{Fsec}{PaintMove()}
    \SetKwFunction{Ft}{MoveTPosition()}
\SetKwProg{Func}{Function}{:}{}
\define{Step\_x , Step\_y}
\initialize{StepperMotors(), ServoMotor(), Sensors(), PaintMove(), Paint()}

\Func{\Fmain (step\_x,step\_y)}
{
\While{ServoMotor()==True}{
Connect to pin 3
}
\While{StepperMotors()==True}{Set speed of step\_x=60 and step\_y=60}}

\Func{\Fsec}
{
\For{ position in (x,y)}{
\State MoveTPosition(x,y);
\State Initiate Painting with Paint();\\}
}
\Func{\Ft}{
\State Calculate Steps for both step\_x and step\_y;\\
}
\Func{Paint()}{
\State Start the paint process by setting Servo position to 90-degree angle\\
\State Delay for 500;\\
\State Set servo position to 0-degree angle for stop
}
\Func{ObstacleDetect()}
{
\If{(Ultrasonic sensor value detect obstacle)}
 {
\State Stop Paint;\\
\State Adjust path for the robot
}
}
\end{algorithm}   
\end{comment}

\begin{figure*}[!htb]
  \centering
  \includegraphics[scale=0.55]{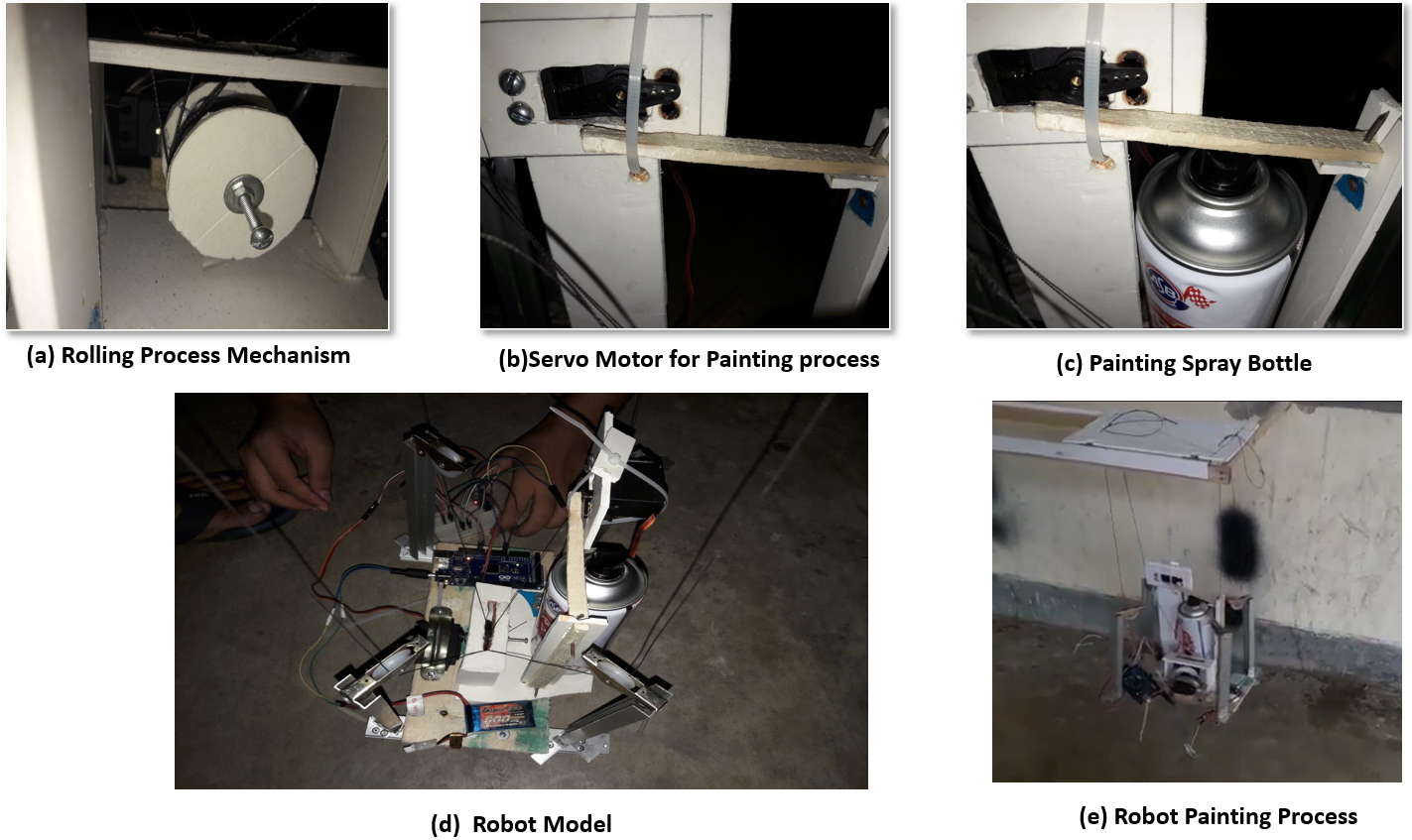}
  \caption{Several Analysing Parts of the Wall Painting Robot}
  \label{fig:method2}
\end{figure*}

\section{Results Analysis and Discussions}
\label{RD}

In order to calculate the steps required to cover a specific distance, first, we calculated the steps per revolution using equation \ref{STR}; using this value, the per-distance steps were calculated using equation \ref{ds}, and finally, using this, the steps for a given distance were measured using equation \ref{d} \cite{liu2024switched}. 
\begin{equation}
\label{STR}
  SPR=\frac{360^{o}}{Step Angle}
\end{equation}

\begin{equation}
\label{ds}
    SPR=\frac{\pi * pulley diameter }{SPR}
\end{equation}

\begin{equation}
\label{d}
    steps=\frac{Distance}{Distance per step}
\end{equation}

In Fig. \ref{fig:method2} represents the total robot wall painting prototype. In Fig. \ref{fig:method2} (a) the rolling process of the robot is shown, (b) shows the pressure enforced by the servo motor to the painting spray machine, and after that, the painting spry bottle is placed as shown in \ref{fig:method2} (c). Again, the total Robot model is shown as in Fig. \ref{fig:method2} (d). Finally, the process of outside wall painting through our proposed model is depicted in \ref{fig:method2} (e). \vspace{2mm}

Although we only painted one square foot, our 1.5 square foot trial was successful. Additionally, the height of the spray prevented it from painting the lower portion of the space. The error is calculated by using the following equation \ref{er}. The highest error obtained from the experimental analysis is 0.5 mm. Further evaluation of the uniform painting coverage overlap ratio is considered by the ratio of the width of the paint stroke with the difference of the width of the paint stroke and the distance between each stroke as shown in equation \ref{or}. In our experimental analysis, we got 10 mm as the width of the paint stroke with 5.5 mm distance. Thus, the overlap ratio was 45\%

\begin{equation}
\label{er}
    Error=\left | Expected_{position}-Actual_{position} \right |
\end{equation}

\begin{equation}
\label{or}
    OR=\frac{Width-Distance}{Width}
\end{equation}
\noindent
\vspace{2mm}

\begin{table}[]
\caption{Evaluation Matrices and Corresponding Values}
\centering
\begin{tabular}{ll}
\hline
\multicolumn{1}{|c|}{\textbf{Evaluation Matrices}} & \multicolumn{1}{c|}{\textbf{Obtained Value}} \\ \hline
\multicolumn{1}{|l|}{Sensor Error Margin} & \multicolumn{1}{l|}{0.02 m}  \\ \hline
\multicolumn{1}{|l|}{Movement Precision}  & \multicolumn{1}{l|}{0.5 mm}  \\ \hline
\multicolumn{1}{|l|}{Overlap Ratio}       & \multicolumn{1}{l|}{45 \%}   \\ \hline
\multicolumn{1}{|l|}{Distance between each Step}   & \multicolumn{1}{l|}{0.01 mm per step}        \\ \hline
\multicolumn{1}{|l|}{Deviation}           & \multicolumn{1}{l|}{2.24 mm} \\ \hline
                     
\end{tabular}
\label{Ev}
\end{table}

\begin{figure}[h]
    \centering
    \includegraphics[scale=0.45]{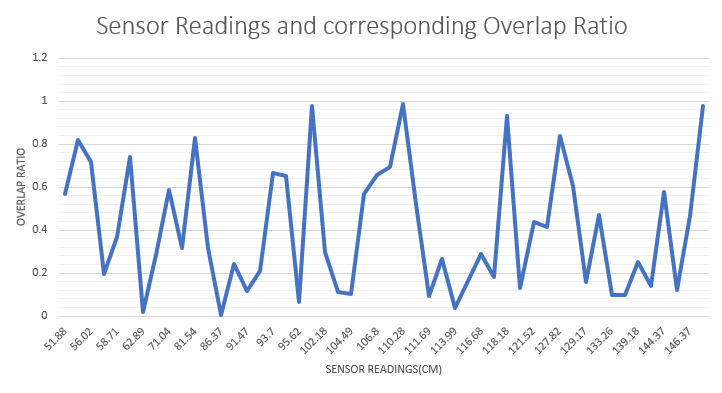}
   \caption{\textcolor{black}{Ultrasonic Sensor Reading and Corresponding Overlap Ratio}}
    \label{fig:f5}
\end{figure}

On the other hand, Fig. \ref{fig:f5} presents the ultrasonic sensor readings and the corresponding overlap ratio achieved. The range of the sensor value is between 50 and 150 cm. The overlap ratio achieved for different sensor values is along the Y-axis. This helps measure the position of the robot with respect to the building wall. Which in turn helps to maintain consistent paint throughout the overall painting procedure.
These readings help determine the robot's position relative to the wall, which is crucial for maintaining consistent paint application \cite{santhosh2024optimizing, rahman2023towards}. When the sensor value is 51.88, it achieves a 57\% overlap ratio, which is considered to be a good ratio to ensure the painting process is thorough. In the case of 86.37 sensor value, the overlap ratio becomes 0, which indicates gaps are painted. Again, with 147.86 and 96.15
the overlap ratio is 98 percent, and 110.28, the ratio is 98\%. This high value of the ratio indicates the multiple coating process without removing the previous coat of paint. When the overlap ratio is low, for instance, 20 or 25 percent and 30 or 31 seconds of time per stroke, we obtain a mediocre paint quality; but, with a 45 percent overlap ratio and between 35 and 36 times per stroke, this quality improved. In table \ref{res}, a different setup is presented for a comparative analysis. 

\begin{table}[h]
\caption{Paint quality measurement based on sensor reading, Overlap ratio, and Time per Stroke}
\centering
\begin{tabular}{|l|l|l|l|l|}
\hline
\multicolumn{1}{|c|}{\textbf{Index}} &
  \multicolumn{1}{c|}{\textbf{Sensor Value}} &
  \multicolumn{1}{c|}{\textbf{OR}} &
  \multicolumn{1}{c|}{\textbf{Stroke Time}} &
  \multicolumn{1}{c|}{\textbf{Quality of Paint}} \\ \hline
1  & 57.1 & 0.2  & 30 & Medium      \\ \hline
2  & 139.18 & 0.25 & 31 & Medium      \\ \hline
3  & 102.18 & 0.3  & 32 & Moderate-High \\ \hline
4 & 76.46 & 0.32  & 32 & Moderate-High \\ \hline
5  & 58.71 & 0.35 & 33 & Moderate-High \\ \hline
6 & 127.42  & 0.4  & 34 & Moderate-High \\ \hline
7  & 121.52 & 0.45 & 35 & High          \\ \hline
8 & 129.17 & 0.45 & 35 & High          \\ \hline
9  & 104.49 & 0.5  & 36 & High          \\ \hline
10  & 104.88  & 0.55 & 38 & High          \\ \hline
\end{tabular}

\label{res}
\end{table}

\section{Conclusion} \label{sec:conclu}
The Remote Control Painting System (RCPS) for painting the outside walls of skyscrapers is a huge step forward in building maintenance technology. By employing robotics, remote control, and cutting-edge materials, RCPS tackles the pressing concerns of risk, effectiveness, and good workmanship in high-rise building painting. 
%With urban landscapes continually changing, such creative fixes will be essential in preserving the aesthetic as well as structural soundness of our tall buildings to ensure that they continue to be considered exponents of contemporary engineering and architecture. 
In this regard, this paper proposes an automatic and robotic solution for exterior painting of high-rise buildings. It presents a prototype with a spray mechanism that moves in four directions using microcontroller-operated servo motors. The system can be upgraded to a robotic system in the future, controlled remotely via a smartphone interface and Bluetooth technology. The system aims to improve productivity and workplace safety. However, the created basic prototype that allows to paint a building's outside walls using a Bluetooth control system. If this system encounters any obstructions, such as a sunroof or an arch, it will not function correctly. This system component needs further development in the future.
%may also be developed.

\ifCLASSOPTIONcaptionsoff
  \newpage
\fi
\bibliographystyle{IEEEtran}
\bibliography{sample}
\end{document}